\documentclass[journal]{IEEEtran}

\ifCLASSINFOpdf
\else
\usepackage[dvips]{graphicx}
\fi
\usepackage{url}
\usepackage{cite}
\usepackage{amsmath,amssymb,amsfonts}
\usepackage{algorithm}
\usepackage{algorithmic}
\usepackage{graphicx}
\usepackage{textcomp}
\usepackage{tikz}
\usepackage{tabularx}

\usetikzlibrary{shadows, arrows.meta, positioning}
\usepackage{xcolor}
\usepackage{array}
\usepackage{caption}
\usepackage{multirow}
\usepackage{booktabs}
\usepackage{hyperref}

\hypersetup{colorlinks=true, citecolor=blue, linkcolor=blue, urlcolor=blue}
\usepackage{orcidlink}
\usepackage{pgfplots}

\pgfplotsset{compat=1.18}
\usepackage{helvet}
\usepackage{sansmath}
\hypersetup{
	colorlinks,
	linkcolor={red!50!black},
	citecolor={blue!50!black},
	urlcolor={blue!80!black}
}

\begin{document}
	
	\title{F-ACVAE: A Federated Adaptive Conditional Variational Auto-Encoder for Privacy-Preserving Intrusion Detection in IoT Networks}
	
	\author{Mohammad Ansarimehr$^{\orcidlink{0009-0004-2864-5617}}$, Somayeh Changiz$^{\orcidlink{0009-0008-7668-9066}}$, Ehsan Baghishani$^{\orcidlink{0009-0002-6849-9357}}$ and Ali Mousavi$^{\orcidlink{0000-0002-4084-6102}}$
		
		\thanks{Corresponding Author: Ali Mousavi (e-mail: mousavi@iau.ac.ir).
			
			All authors are with Department of Computer Engineering, Ne. C., Islamic Azad University, Neyshabur, Iran (e-mail: mohamad.ansarimehr@iau.ac.ir, s.changiz@iau.ac.ir, ehsan.baghishani@iau.ac.ir, mousavi@iau.ac.ir).}}
	
	\markboth{IEEE Transactions on Network and Service Management,~Vol.~XX, No.~X, 2025}%
	{Mohammad Ansarimehr \MakeLowercase{\textit{et al.}}: F-ACVAE}
	
	\maketitle
	
	\begin{abstract}
		The rapid proliferation of Internet of things (IoT) devices has significantly expanded the cyber-attack surface, necessitating robust and privacy-preserving intrusion detection systems (IDS). However, centralized learning approaches often suffer from severe performance degradation due to high-dimensional traffic data, extreme class imbalance, and highly non-independent and identically distributed (non-IID) data across heterogeneous edge devices.
		To address these challenges, this paper proposes F-ACVAE, a federated adaptive conditional variational autoencoder framework that enables collaborative model training across distributed IoT devices without sharing raw data. F-ACVAE incorporates selective parameter aggregation, where local encoders remain private while globally shared components are synchronized to preserve discriminative latent structures. To further enhance stability under extreme non-IID settings and feature distribution shifts, we introduce a novel constrained momentum Gaussian aggregation (CMGA) strategy that combines update clamping with momentum-based smoothing to mitigate client drift.
		Extensive experiments on the N-BaIoT dataset demonstrate that F-ACVAE achieves an average accuracy and macro F1-score of 99\%, outperforming state-of-the-art baselines. Moreover, the selective aggregation mechanism reduces communication overhead by approximately 62\%, making the framework particularly suitable for resource-constrained IoT environments. These results highlight the effectiveness of F-ACVAE in achieving high detection performance while ensuring privacy preservation and communication efficiency.
	\end{abstract}
	
	
	\IEEEpeerreviewmaketitle 
	
	\section{Introduction}
	\label{sec:introduction}
	
	{I}{nternet} of things (IoT) devices are now widespread in smart homes, industry, healthcare, and urban infrastructure.
	This rapid expansion has significantly increased the cyberattack surface for billions of resource-constrained devices, as evidenced by recent security benchmarks \cite{meidan2018nbaiot}.
	Large-scale botnet attacks, such as Mirai \cite{kolias2017mirai} and Bashlite \cite{anton2018mirai}, have demonstrated how millions of infected IoT devices can be orchestrated to launch massive distributed denial-of-service (DDoS) attacks.
	Traditional centralized intrusion detection systems (IDS) require the transmission of raw traffic data to a central server for analysis.
	However, this approach presents three critical challenges in IoT networks.
	First, the exposure of sensitive raw data raises severe privacy concerns, potentially violating international regulations such as the general data protection regulation (GDPR) \cite{mothukuri2021federated} and the California consumer privacy act (CCPA).
	Second, continuous data transmission imposes significant communication overhead on bandwidth-limited networks.
	Third, centralized models often demonstrate poor performance due to highly non-IID data distributions across devices and severe class imbalance in network traffic \cite{dinh2024}.
	
	Among deep learning approaches for IoT intrusion detection, existing methods are generally divided into two main categories: traditional supervised machine learning models and autoencoder-based generative models.
	Supervised approaches primarily depend on classifiers trained on labeled traffic data.
	Early studies utilized classical algorithms such as random forest (RF), gradient boosting (XGBoost), and support vector machines (SVM).
	These models achieved high accuracy on benchmarks like CIC-IoT and N-BaIoT.
	Recent works have employed deep supervised architectures, such as CNN-LSTM hybrids and BiLSTM with attention mechanisms, as well as ensemble defenses enhanced by adversarial training.
	Such models effectively capture spatial and temporal patterns in network flows, consistently reporting accuracies exceeding 99\% \cite{deshmukh2025} and demonstrating strong robustness against known and adversarial attacks.
	However, these models require the centralized collection of raw labeled data across devices, which raises significant privacy concerns and causes substantial communication overhead in resource-constrained IoT environments.
	
	In contrast, autoencoder-based methods employ an unsupervised or semi-supervised paradigm, learning compact representations of normal traffic and detecting anomalies through high reconstruction errors.
	Initial efforts focused on vanilla and denoising autoencoders, while later advances incorporated variational autoencoders (VAEs) and sparse variants \cite{ain2025}, as well as convolutional backbones to better model complex feature distributions \cite{padmavathi2025}.
	Such generative models are particularly effective under severe class imbalance and do not require balanced attack labels.
	To overcome the limitations of both categories, the constrained twin variational autoencoder (CTVAE) has been proposed as an effective approach \cite{dinh2024}.
	It utilizes class-conditioned Gaussian priors and employs a unique hermaphrodite regularization path, enforcing strong structural separation between benign and malicious latent representations.
	
	The CTVAE method effectively identifies complex attacks; however, its design remains centralized.
	This architecture requires all raw data from thousands of IoT devices to be sent to a single central server, which seriously violates data privacy and security regulations, a concern particularly critical when devices handle sensitive information \cite{mothukuri2021federated}.
	Furthermore, the continuous transmission of massive raw data in centralized intrusion detection frameworks imposes heavy communication overhead that often exceeds the capacity of resource-constrained IoT devices, motivating the adoption of federated learning paradigms \cite{hernandez2025systematic}.
	To address these challenges, federated learning (FL) has emerged as an effective solution, enabling the collaborative training of a robust global model by sharing only model updates, such as weights, instead of private local data.
	Consequently, the privacy of each device is maintained, and bandwidth requirements are significantly reduced, making FL well-suited for distributed IoT networks.
	As a privacy-by-design paradigm, FL allows models to be trained across distributed devices without sharing raw data.
	Standard FL methods, including FedAvg and FedProx for non-IID data \cite{sahu2018noniid}, along with their modern variants, are widely applied to IoT intrusion detection systems to preserve privacy while maintaining accuracy \cite{li2025}.
	However, these algorithms face a major limitation: they suffer from significant accuracy degradation when data is highly non-IID, including extreme label skew and feature distribution shifts prevalent in real-world IoT deployments.
	
	In this paper, we propose F-ACVAE, a federated adaptive conditional variational autoencoder for privacy-preserving intrusion detection in highly heterogeneous IoT networks. 
	F-ACVAE addresses the limitations of traditional federated learning on complex VAE models through selective knowledge sharing: local encoders remain strictly private on each device, while only essential shared components—the decoder and adaptive conditional modules—are aggregated. 
	This selective aggregation substantially reduces communication overhead while preserving data privacy.
	To handle extreme non-IID data, including label skew and feature shifts, we introduce constrained momentum Gaussian aggregation (CMGA). 
	CMGA employs momentum-based smoothing to stabilize global updates and enforce class-wise separation in the latent space, preventing overlap between normal and attack distributions and reducing false positives. 
	By preserving a fixed global model structure and aggregating only informative weights, F-ACVAE minimizes communication cost without compromising detection performance.
	
	The primary contributions of this work are summarized as follows:
	\begin{itemize}
		\item We propose F-ACVAE, a novel federated adaptive conditional variational autoencoder that enables privacy-preserving intrusion detection in highly heterogeneous IoT environments.
		
		\item We introduce CMGA, a constrained momentum-based aggregation strategy that stabilizes global updates and preserves class-wise latent space separability under extreme non-IID data distributions.
		
		\item We develop a selective federation protocol that explicitly decouples private and shared knowledge: local encoders remain strictly private on edge devices, while only globally meaningful components (the decoder and adaptive conditional modules) are aggregated, resulting in strong privacy guarantees and approximately 62\% reduction in communication overhead.
		
		\item We demonstrate through extensive experiments on the N-BaIoT dataset that F-ACVAE consistently achieves an average accuracy and macro F1-score of 99\%, significantly outperforming state-of-the-art baselines.
	\end{itemize}
	
	The remainder of this paper is organized as follows: Section~\ref{sec:related} reviews the related literature on IoT security and federated learning.
	Section~\ref{sec:background} provides the necessary technical background on the variational autoencoder and the Flower federated learning framework.
	Section~\ref{sec:method} details the proposed F-ACVAE framework and the CMGA strategy.
	Section~\ref{sec:experiments} presents the experimental setup, while Section~\ref{sec:results} provides a comprehensive analysis of the results.
	Section~\ref{sec:future} discusses potential future research directions, and finally, Section~\ref{sec:conclusion} concludes the paper.
	
	\section{Related Work}
	\label{sec:related}
	
	This section reviews the relevant literature by first analyzing centralized intrusion detection systems (IDS).
	These systems are categorized into traditional machine learning and autoencoder-based approaches.
	Furthermore, this section discusses the evolution and challenges of federated learning (FL), providing the necessary context for understanding privacy-preserving IDS in heterogeneous Internet of things (IoT) settings.
	
	\subsection{Traditional Machine Learning Approaches}
	IoT intrusion detection systems (IDS) initially relied on signature-based methods \cite{buczak2016survey, hajiheidari2019ids}, which effectively identified known attack patterns; however, they failed against zero-day and polymorphic threats \cite{sommer2010outside}, as prevalent in modern IoT botnets such as Mirai and Bashlite \cite{meidan2018nbaiot, dhakal2024enhancing}.
	While anomaly-based methods were introduced to improve adaptability \cite{khraisat2019survey}, they often produced high false-positive rates in dynamic environments.
	To address these limitations, machine learning (ML) and deep learning (DL) models have dominated recent literature \cite{algaradi2020survey, abdelmoumin2022ml, yang2022mthids}, particularly for designing adaptive and intelligent intrusion detection systems \cite{hussain2020ml} that can generalize to evolving threats in IoT environments \cite{xin2018ml}.
	Early IDS utilized classical ML algorithms, including random forest (RF) and support vector machines (SVM) \cite{tvt_ml_2022, hasan2014svmrf}, as well as gradient boosting (XGBoost) methods \cite{buczak2016survey, bhati2021xgb}, which often relied on extensive manual feature engineering to achieve high accuracy on benchmark datasets such as CIC‑IoT \cite{dekeersmaeker2023survey} and UNSW‑NB15 \cite{moustafa2015unsw}.
	Recent research has shifted significantly toward DL models \cite{hussain2020ml, shone2018dl, abdulhammed2019dl} to overcome the limitations of manual feature engineering \cite{meidan2018nbaiot}.
	For instance, convolutional neural networks (CNNs) have been widely adopted to automatically extract spatial features from network traffic representations and packet headers \cite{popoola2021hybrid}.
	More recently, graph-convolutional approaches have been proposed to capture complex topological relationships in vehicular IoT traffic \cite{deng2023flow}, while other architectures utilize recurrent neural networks (RNNs) \cite{yin2017dl} and deep autoencoder-based models \cite{shone2018dl} for robust detection.
	Similarly, long short-term memory (LSTM) networks and their variants focus on capturing temporal dependencies in sequence-based data \cite{sherstinsky2020fundamentals, popoola2021hybrid, ullah2021deeplearning}.
	Furthermore, hybrid deep architectures combine CNNs with BiLSTM or integrate attention mechanisms to exploit multi-dimensional correlations \cite{popoola2021hybrid, vu2020transfer, yang2022mthids}, with advanced frameworks reporting AUC scores above 99\% on benchmark datasets.
	Despite their strong performance, these approaches share fundamental limitations, including the requirement for large volumes of labeled training data and a strong dependence on centralized data aggregation \cite{buczak2016survey, shone2018dl}.
	This centralization introduces two critical issues: first, it leads to severe privacy violations, often conflicting with data protection regulations such as the general data protection regulation (GDPR) and the California consumer privacy act (CCPA); and second, it creates high communication overhead, making such models unsuitable for privacy-sensitive and resource-constrained IoT environments \cite{li2020federated, badii2020smartcity}.
	These limitations have motivated a growing body of research on federated and other decentralized collaborative learning paradigms, which aim to reduce communication burden and keep data local while preserving model performance on distributed devices such as those in IoT systems \cite{li2020federated, sahu2018noniid}.
	
	\subsection{Generative Learning and Autoencoder-Based Approaches}
	Unsupervised and semi-supervised autoencoders (AEs) have emerged as highly effective paradigms for IoT anomaly detection by learning compact latent representations of benign traffic and reconstructing normal patterns within a low-dimensional manifold \cite{ferrag2020survey, meidan2018nbaiot}.
	These models identify potential cyber-attacks by monitoring high reconstruction errors, which indicate deviations from the learned normal manifold of network behavior \cite{sakurada2014anomaly, chalapathy2019deep, meidan2018nbaiot}.
	While initial research focused on deterministic architectures such as denoising and sparse AEs to handle noisy IoT data \cite{langarica2023contrastive}, the field has gradually and significantly shifted toward variational autoencoders (VAEs) \cite{kingma2014auto}.
	VAEs introduce a probabilistic framework that regularizes the latent space toward a simple Gaussian prior, enabling more continuous and robust data representations \cite{kingma2014auto}; however, multiple studies have shown that such regularization alone often fails to enforce clear discriminative separation between normal and anomalous traffic clusters \cite{an2015variational, ding2021vae}.
	To address this, twin variational autoencoders (TVAEs) were introduced to capture distinct latent distributions for benign and malicious traffic separately, leading to the development of the constrained twin variational autoencoder (CTVAE) which currently holds the state-of-the-art position for the N-BaIoT dataset \cite{dinh2024}.
	CTVAE achieves superior detection performance by leveraging class-conditioned Gaussian priors to model benign and malicious traffic separately, and by using the hermaphrodite component together with an additional constraint term to increase the separation between latent clusters, thus preserving discriminative structure under class imbalance \cite{dinh2024}.
	Despite these algorithmic advances, existing CTVAE implementations rely exclusively on centralized training, which necessitates aggregating raw IoT traffic on a single server. This centralization exposes sensitive network metadata to significant privacy risks and may conflict with modern data protection regulations such as GDPR and CCPA, thereby motivating privacy preserving decentralized learning solutions \cite{li2020federated, khraisat2019survey, badii2020smartcity}.
	
	\subsection{Federated Learning and Decentralized Approaches}
	The urgent need for data privacy and reduced network overhead has accelerated the adoption of federated learning (FL) \cite{li2020federated, lim2021federated}, which has quickly become the leading framework for privacy-preserving collaborative training.
	FL allows many edge devices to collaboratively train a high-quality global model by sharing only model weight updates instead of raw data \cite{li2020federated}, with comprehensive surveys further detailing the underlying strategies, challenges, and future research directions of the field \cite{lim2021federated}.
	Recent studies investigate architectural patterns and design considerations critical for building robust FL systems \cite{lim2021federated}.
	Foundational works such as DIoT \cite{nguyen2019diot} and federated malware detection systems \cite{rey2021federated} have demonstrated that FedAvg and its variants can achieve accuracy comparable to centralized training on IoT malware and intrusion detection datasets.
	Subsequent research has expanded the scope of FL-based IDS by proposing communication- and resource-efficient frameworks such as FD-IDS, which integrate federated learning with knowledge distillation \cite{peng2025}.
	Personalized FL variants, including APFed and DFF-FL, have been introduced to better handle data heterogeneity, while knowledge-distillation-enhanced training in FD-IDS further improves model performance \cite{li2025, peng2025, almansour2025adaptive}.
	Further studies conducted large-scale emulations using CNN-BiLSTM backbones to evaluate FL-based IDS performance \cite{begum2024bflids, albanbay2025federated, baidar2025hybrid}.
	Other works focused on enhancing intrusion detection in IoT networks through federated learning, including specialized applications such as the FELIDS framework for agricultural IoT \cite{friha2022}.
	In particular, some studies employed generative models, such as the federated variational autoencoder (FedVAE), to enhance data privacy \cite{cui2023fedvae}.
	Other works focused on improving security and robustness by integrating differential privacy \cite{ruzafaalcazar2023}, homomorphic encryption in some FL IDS designs \cite{wang2024nids, vyas2024survey}, or blockchain-secured updates, exemplified by BFLIDS \cite{begum2024bflids}.
	Additionally, hybrid FL systems have been developed to enhance zero-day intrusion detection in decentralized IoT environments by combining GANs for class-balancing augmentation and implementing proactive detection pipelines \cite{tabassum2022fedgan}.
	Real-device experiments on Raspberry Pi clusters have validated the scalability and convergence of federated learning under non-IID conditions \cite{peng2025}.
	However, although FL preserves raw data privacy, its performance often declines significantly when faced with the data heterogeneity common in real-world IoT environments \cite{sahu2018noniid}.
	Specifically, the non‑IID nature of network traffic, with extreme label skew, has been shown to compromise the convergence of the aggregated global model \cite{li2020federated}, while feature distribution shifts further degrade its generalization in federated learning \cite{zhu2021noniid}.
	To address this challenge, specialized federated learning strategies are required that can better cope with heterogeneous client distributions while preserving model integrity and discriminative structure in decentralized settings \cite{sattler2020clustered}.
	
	\section{Background}
	\label{sec:background}
	
	This section describes the two fundamental components necessary for implementing the proposed framework: the variational autoencoder (VAE) and the Flower federated learning framework.
	
	\subsection{Variational Autoencoder (VAE) Framework}
	The VAE \cite{kingma2014auto} is a fundamental probabilistic generative model that maps high-dimensional input data $x \in \mathbb{R}^{d_x}$ into a compact latent representation $z \in \mathbb{R}^{d_z}$. 
	Unlike standard autoencoders, the VAE introduces a probabilistic formulation by assuming a prior distribution over the latent variables, typically modeled as an isotropic Gaussian $p(z) = \mathcal{N}(0, I)$.
	
\begin{figure}[htbp]
		\centering
		\includegraphics[width=0.45\textwidth]{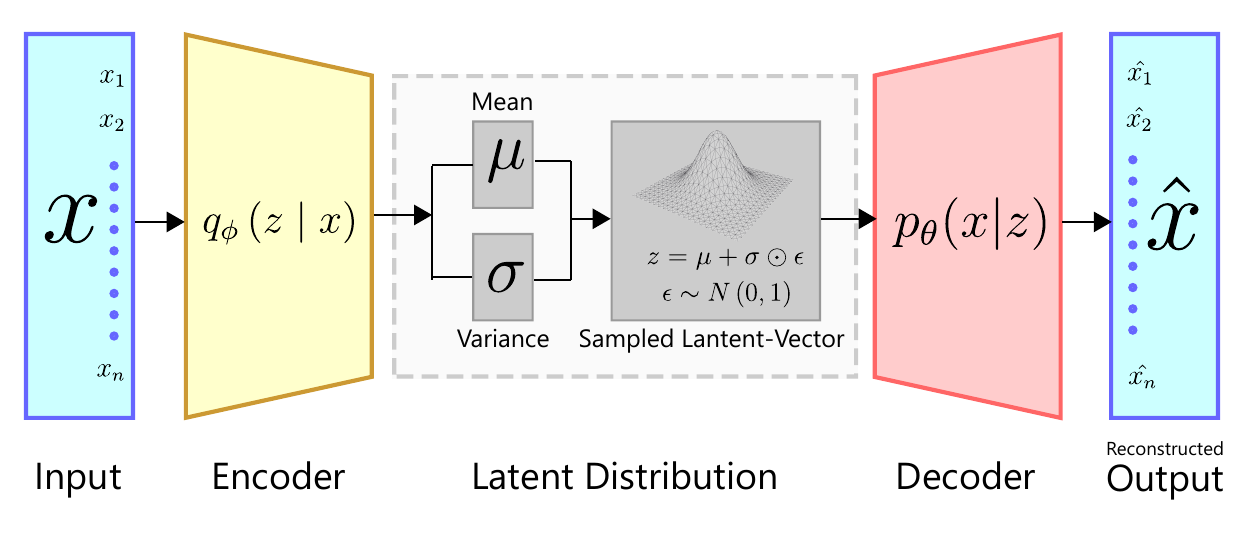}
		\caption{Standard architecture of VAE consisting of an encoder, a latent space with Gaussian prior, and a decoder.}
		\label{fig:vae}
	\end{figure}

	As shown in Figure~\ref{fig:vae}, a VAE consists of two main components:
	\begin{enumerate}
		\item Inference Network (Encoder): 
		The encoder, parameterized by $\phi$, approximates the true posterior $p(z|x)$, which is not directly computable, with a variational distribution $q_\phi(z|x)$.
		It outputs the mean vector $\mu(x)$ and the log-variance vector $\log \sigma^2(x)$ of a Gaussian distribution:
		\begin{equation}
			q_\phi(z|x) = \mathcal{N}\big(z \mid \mu(x), \mathrm{diag}(\sigma^2(x))\big).
		\end{equation}
		To enable gradient-based optimization, the reparameterization trick is applied:
		\begin{equation}
			z = \mu(x) + \sigma(x) \odot \epsilon, \quad \epsilon \sim \mathcal{N}(0, I).
		\end{equation}
		
		\item Generative Network (Decoder):  
		The decoder, parameterized by $\theta$, defines the conditional likelihood $p_\theta(x|z)$ and reconstructs the input from the latent variable.
		The reconstructed output is sampled (or taken as the mean) from this distribution:
		\begin{equation}
			\hat{x} \sim p_\theta(x|z) \;\; \text{or} \;\; \hat{x} = \mathbb{E}_{p_\theta(x|z)}[x], \quad \hat{x} \in \mathbb{R}^{d_x}.
		\end{equation}
	\end{enumerate}
	
	The VAE is trained by minimizing the negative Evidence Lower Bound (ELBO), which is formulated as the sum of a reconstruction term and a regularization term:
	\begin{equation}
		\mathcal{L}_{\text{VAE}} = \mathcal{L}_{\text{Recon}} + \mathcal{L}_{\text{KL}}
	\end{equation}
	where
	\begin{equation}
		\begin{aligned}
			\mathcal{L}_{\text{Recon}} &= -\mathbb{E}_{q_\phi(z|x)} \left[ \log p_\theta(x|z) \right], \\[4pt]
			\mathcal{L}_{\text{KL}} &= \beta \, D_{\text{KL}}\big( q_\phi(z|x) \,\|\, p(z) \big).
		\end{aligned}
	\end{equation}
	
	The reconstruction loss encourages accurate data reconstruction, while the Kullback--Leibler (KL) divergence regularizes the latent space by enforcing similarity between the approximate posterior distribution $q_\phi(z|x)$ with the prior distribution $p(z)$.
	
	The hyperparameter $\beta$ regulates the trade-off between the KL divergence term and the reconstruction loss.
	When $\beta = 1$, the model reduces to the standard VAE formulation, whereas larger values of $\beta$ impose stronger regularization on the latent space.
	
	In intrusion detection, increasing $\beta$ enhances the regularization of the latent space, leading to more compact representations of normal traffic. 
	Anomaly detection is subsequently performed by computing the reconstruction error of a test sample $x$, defined as:
	\begin{equation}
		s(x) = \|x - \hat{x}\|_2^2, \quad x, \hat{x} \in \mathbb{R}^{d_x}.
	\end{equation}
	A test sample $x$ is classified as an intrusion if its reconstruction error exceeds the threshold $\tau$:
	\begin{equation}
		s(x) > \tau.
	\end{equation}
	
	The training procedure of the VAE, based on stochastic gradient descent, is detailed in Algorithm~\ref{alg:vae}.
	
	\begin{algorithm}[htbp]
		\caption{Stochastic Gradient Descent Training of VAE}
		\label{alg:vae}
		\begin{algorithmic}[1]
			\STATE \textbf{Input:} Training dataset $\mathcal{X}$, learning rate $\eta$, mini-batch size $M$, KL weight $\beta$
			\STATE \textbf{Output:} Optimized encoder parameters $\phi$ and decoder parameters $\theta$
			\STATE \textbf{Initialize:} $\phi$ and $\theta$ randomly
			\WHILE{not converged}
			\STATE Sample a mini-batch $\{x^{(i)}\}_{i=1}^M$ from $\mathcal{X}$
			\STATE Compute mean and log-variance using encoder: $$ (\mu^{(i)}, \log \sigma^{2(i)}) \gets q_\phi(z \mid x^{(i)}) $$
			\STATE Sample noise: $\epsilon^{(i)} \sim \mathcal{N}(0, I)$ for each $i=1,\dots,M$
			\STATE Reparameterization: $z^{(i)} \gets \mu^{(i)} + \sigma^{(i)} \odot \epsilon^{(i)}$
			\STATE Reconstruct using decoder: $\hat{x}^{(i)} \gets \mathbb{E}_{p_\theta(x|z^{(i)})}[x]$
			\STATE Compute reconstruction loss for the mini-batch:
			\[
			\mathcal{L}_{\text{Recon}}(\phi, \theta) = -\frac{1}{M} \sum_{i=1}^M \mathbb{E}_{q_\phi(z|x^{(i)})} [\log p_\theta(x^{(i)} \mid z^{(i)})]
			\]
			\STATE Compute KL divergence loss for the mini-batch:
			\[
			\mathcal{L}_{\text{KL}}(\phi) = \frac{1}{M} \sum_{i=1}^M D_{\text{KL}}(q_\phi(z \mid x^{(i)}) \| p(z))
			\]
			\STATE Compute total loss: $\mathcal{L}(\phi, \theta) = \mathcal{L}_{\text{Recon}}(\theta) + \beta \mathcal{L}_{\text{KL}}(\phi)$
			\STATE Update encoder: $\phi \gets \phi - \eta \nabla_\phi \mathcal{L}$
			\STATE Update decoder: $\theta \gets \theta - \eta \nabla_\theta \mathcal{L}$
			\ENDWHILE
		\end{algorithmic}
	\end{algorithm}
	
	\subsection{Flower Federated Learning Framework}
\begin{figure}[htbp]
		\centering
		\includegraphics[width=0.92\columnwidth]{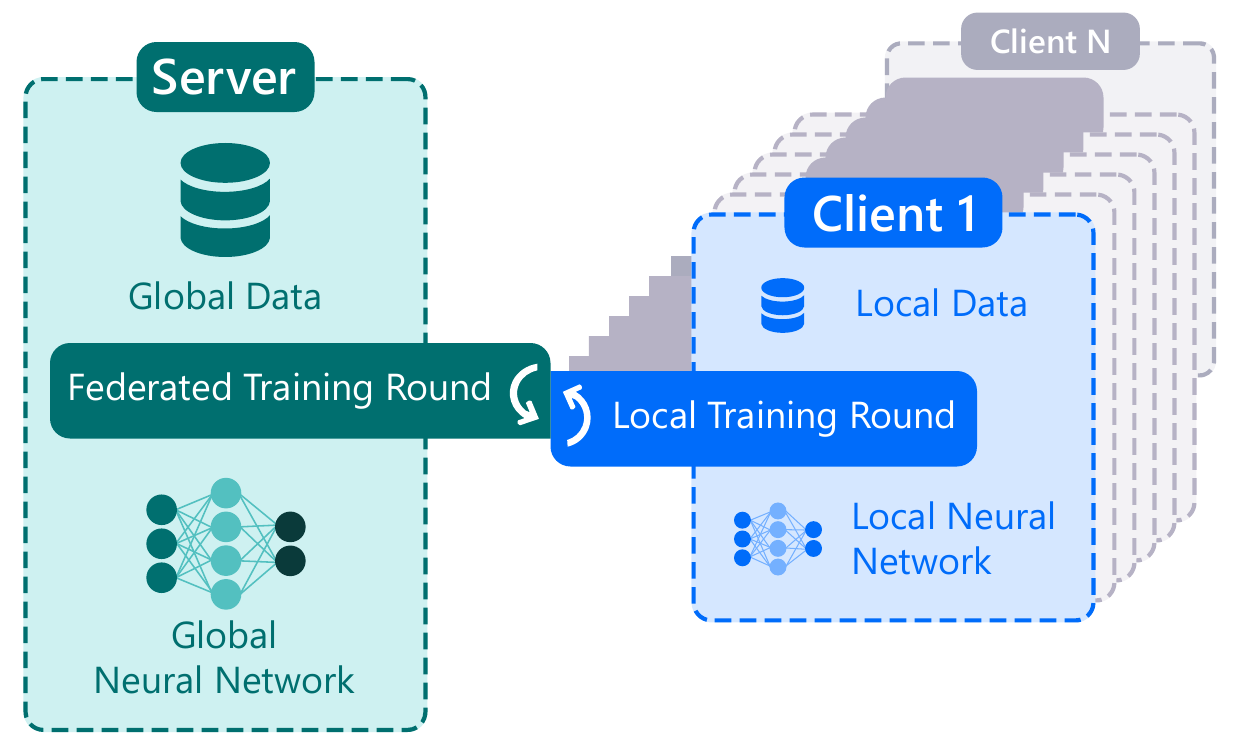}
		\caption{Flower federated learning architecture \cite{beutel2024flower}.}
		\label{fig:flower}
	\end{figure}

	The Flower (flwr) framework \cite{beutel2024flower} serves as the primary engine for coordinating the F-ACVAE system, handling communication between the central server ($S$) and diverse IoT clients ($\mathcal{C}$), as shown in Figure~\ref{fig:flower}.
	Flower's modularity facilitates the implementation of custom aggregation strategies, a capability critical for embedding our novel constrained momentum Gaussian aggregation (CMGA) strategy.
	The framework coordinates the federated process through discrete communication rounds ($t$) managed in three core steps.
	
	\begin{enumerate}
		\item Distribution Phase: The server selects a subset of clients ($\mathcal{S}_t$) and transmits the global model parameters ($\theta_G^t$) and configuration settings ($\text{config}_t$):
		\begin{equation}
			\text{Call}(\mathcal{S}_t) = \text{Fit}(\theta_G^t, \text{config}_t).
		\end{equation}
		
		\item Local Training Phase: Each client $k \in \mathcal{S}_t$ trains the model on its private dataset $\mathcal{D}_k$ by minimizing the adaptive conditional VAE loss function $\mathcal{L}_{\text{ACVAE}}$ to obtain updated local parameters:
		\begin{equation}
			\theta_k^{t+1} \leftarrow \arg\min_{\theta_k} \mathcal{L}_{\text{ACVAE}}(\theta_k ; \mathcal{D}_k).
		\end{equation}
		Each client then returns the update difference $\Delta \theta_k = \theta_k^{t+1} - \theta_G^t$ and its local sample count $n_k$ to the server.
		Here, $K$ denotes the total number of clients in the federated learning system, and each $\mathcal{D}_k$ represents the local dataset held by client $k$.
		
		\item Custom Aggregation: The server invokes the CMGA strategy via the \texttt{aggregate\_fit} function to compute the new global parameters:
		\begin{equation}
			\theta_G^{t+1} = \text{Aggregate}_{\text{CMGA}}(\{(\Delta \theta_k, n_k)\}_{k \in \mathcal{S}_t}, \theta_G^t).
		\end{equation}
		This flow enables F-ACVAE to replace standard FedAvg with our proposed CMGA aggregation logic, which is essential for preserving the conditioned latent structure under severe non-IID conditions and heterogeneous edge environments. 
	\end{enumerate}
	
	The base federated round in Flower is shown in Algorithm~\ref{alg:flower-base}, while the CMGA-enhanced process is detailed in Algorithm~\ref{alg:flower-cmga}.
	
	\begin{algorithm}[htbp]
		\caption{Flower-Based Federated Learning Server}
		\label{alg:flower-base}
		\begin{algorithmic}[1]
			\STATE \textbf{Input:} Initial global parameters $\theta_G^0$, number of federated rounds $T$, aggregation strategy $S$, client datasets $\{\mathcal{D}_k\}_{k=1}^K$ where $K$ is the total number of clients
			\STATE \textbf{Output:} Final global model $\theta_G^T$
			\STATE \textbf{Initialize:} $\theta_G \gets \theta_G^0$
			\FOR{$t = 1$ \textbf{to} $T$}
			\STATE \textbf{Client Selection:} $S$ selects a subset of clients $\mathcal{S}_t$
			\STATE \textbf{Configuration:} $S$ sends current global model $\theta_G^{t-1}$ and configuration to each client $k \in \mathcal{S}_t$
			\FOR{each client $k \in \mathcal{S}_t$ \textbf{in parallel}}
			\STATE \textbf{Client Execution ($k$):}
			\STATE (a) Compute $\Delta \theta_k \gets \text{LocalTrain}(\theta_G^{t-1}, \mathcal{D}_k)$
			\STATE (b) Send $(\Delta \theta_k, n_k)$ back to server $S$
			\ENDFOR
			\STATE \textbf{Global Aggregation:}
			$$ \theta_G^t \gets S.\text{aggregate\_fit}(\{(\Delta \theta_k, n_k)\}_{k \in \mathcal{S}_t}) $$
			\STATE \textbf{Evaluation:} $S$ evaluates the global model $\theta_G^t$
			\ENDFOR
			\STATE \textbf{Return} $\theta_G^T$
		\end{algorithmic}
	\end{algorithm}
	
	\section{Proposed Methodology}
	\label{sec:method}
	
	This section presents the federated adaptive conditional variational autoencoder (F-ACVAE), a framework designed for robust and privacy-preserving intrusion detection in decentralized IoT environments, as illustrated in Figure~\ref{fig:flower}. 
	Unlike centralized learning approaches, which often exhibit instability under severe non-IID data distributions, F-ACVAE enables collaborative intrusion detection across heterogeneous IoT devices through the following five fundamental components:
	
	\begin{enumerate}
		\item {Adaptive conditional VAE:} Learns discriminative latent representations by incorporating class-conditioned information into the variational autoencoder framework.
		
		\item {Constrained momentum Gaussian aggregation (CMGA):} Stabilizes global model updates through selective parameter aggregation and momentum-based smoothing.
		
		\item {Local training strategy:} Ensures client-side autonomy and confidentiality by keeping raw data strictly local during optimization.
		
		\item {Momentum-stabilized aggregation (MSA):} Reduces model drift under non-IID settings by leveraging historical update information.
		
		\item {Hybrid latent feature classification:} Integrates generative latent representations with ensemble-based discriminative classification.
	\end{enumerate}
	
		
		
		
		

	\subsection{Adaptive Conditional VAE Architecture (ACVAE)}
	Let $x_i \in \mathbb{R}^{d_x}$ denote an input feature vector and $y_i \in \{1,\dots,C\}$ its associated class label.
	The encoder network $q_\phi$ maps the input sample to a stochastic latent representation parameterized by a Gaussian posterior distribution:
	\begin{equation}
		q_\phi(z \mid x) = \mathcal{N}\big(z \mid \mu_\phi(x), \mathrm{diag}(\sigma_\phi^2(x))\big)
	\end{equation}
	where $\mu_\phi(x)$ and $\sigma_\phi(x)$ denote the encoder outputs.
	A latent variable is sampled using the reparameterization trick:
	\begin{equation}
		z = \mu_\phi(x) + \sigma_\phi(x) \odot \epsilon, \quad \epsilon \sim \mathcal{N}(0, I)
	\end{equation}
	which enables end-to-end backpropagation through stochastic layers.
	
	To incorporate class-dependent information while maintaining computational efficiency, the proposed ACVAE introduces a learnable class embedding layer $C_\theta(\cdot)$ that maps each label $y$ to a low-dimensional embedding vector $C_y \in \mathbb{R}^L$.
	The final conditioned latent representation is obtained via additive embedding injection:
	\begin{equation}
		z_{\mathrm{cond}} = z + \lambda_c \, C_y
	\end{equation}
	where $\lambda_c > 0$ is a class-conditioning scaling coefficient that controls the influence of label-specific information on the latent representation. Unlike a standard CVAE that conditions the inference network on $y$, ACVAE injects label information directly into the sampled latent code through additive embedding, yielding an equivalent conditional generative mapping $p_\theta(x|z+ \lambda_cC_y)$.
	The embedding dimension $L$ is chosen to match the latent dimensionality, ensuring compatibility for additive conditioning.
	
	This operation adaptively shifts latent representations associated with different classes, thereby enhancing separability between benign and malicious traffic patterns without introducing additional probabilistic constraints or complex architectural components.
	The conditioned latent vector $z_{\mathrm{cond}}$ is subsequently passed to the decoder network $p_\theta$ to reconstruct the input:
	\begin{equation}
		\hat{x} = p_\theta(z_{\mathrm{cond}}).
	\end{equation}
	The complete adaptive conditional latent generation procedure is detailed in Algorithm~\ref{alg:acvae}.
	
	\begin{algorithm}[htbp]
		\caption{Adaptive Conditional Latent Generation}
		\label{alg:acvae}
		\begin{algorithmic}[1]
			\STATE \textbf{Input:} Feature vector $x$, class label $y$
			\STATE \textbf{Output:} Conditioned latent representation $z_{\mathrm{cond}}$
			\STATE \textbf{Initialize:} Encoder $q_\phi$, class embedding network $C_\theta$, scaling coefficient $\lambda_c$
			\STATE Compute encoder parameters: $(\mu, \sigma) \gets q_\phi(x)$
			\STATE Sample noise: $\epsilon \sim \mathcal{N}(0, I)$
			\STATE Compute latent vector: $z \gets \mu + \epsilon \odot \sigma$
			\STATE Compute class embedding: $C_y \gets C_\theta(y)$
			\STATE Condition latent vector: $z_{\mathrm{cond}} \gets z + \lambda_c \times C_y$
			\STATE \textbf{Return} $z_{\mathrm{cond}}$
		\end{algorithmic}
	\end{algorithm}
	
	\subsection{Constrained Momentum Gaussian Aggregation (CMGA)}
	\label{sec:cmga}
	The core of the F-ACVAE framework is the CMGA strategy, which coordinates the synchronization of model parameters while preserving local discriminative structures.
	Unlike standard federated averaging (FedAvg), CMGA introduces a momentum term to stabilize the updates of the shared latent space and class-specific Gaussian priors.
	
	In each communication round $t$, the central server receives the locally updated mappers, decoders, and Gaussian priors from the participating clients.
	To mitigate the effects of extreme non-IID data and prevent latent space collapse, the global parameters $\theta_{G}^{t}$ are updated using a momentum-based moving average:
	
	\begin{equation}
		\theta_{G}^{t} = (1 - \alpha) \theta_{G}^{t-1} + \alpha \left( \frac{1}{K} \sum_{k=1}^{K} \theta_{k}^{t} \right)
	\end{equation}
	where $\alpha \in [0, 1]$ is the aggregation momentum coefficient, and $\theta_{k}^{t}$ represents the subset of parameters (specifically the mapper, decoder, and class priors) uploaded by client $k$.
	In our implementation, $\alpha$ is set to 0.1, ensuring that the global model retains 90\% of its previous structural knowledge while incrementally incorporating local updates.
	This smoothing mechanism significantly stabilizes convergence across heterogeneous IoT devices.
	
	\begin{algorithm}[htbp]
		\caption{Constrained Momentum Gaussian Aggregation}
		\label{alg:flower-cmga}
		\begin{algorithmic}[1]
			\STATE \textbf{Input:} Number of clients $K$, number of rounds $T$, local epochs $E$, momentum coefficient $\alpha$, learning rate $\eta$
			\STATE \textbf{Output:} Global shared parameters $\theta_{G}$ (Mapper, Decoder, Class-Specific Gaussian Priors)
			\STATE \textbf{Initialize:} $\theta_{G}^0 \gets$ random initialization
			\FOR{each round $t = 1$ to $T$}
			\FOR{each client $k \in K$ \textbf{in parallel}}
			\STATE \textbf{Local Update:} $\Delta \theta_k \gets \text{Algorithm~\ref{alg:client}}(\mathcal{D}_k, \theta_G^{t-1}, E, \eta)$
			\STATE \textbf{Extract Shared Parameters:} $\theta_k^t \subset \Delta \theta_k$
			\STATE \textbf{Send to Server:} $\theta_k^t$
			\ENDFOR
			\STATE \textbf{Global Aggregation:} $\bar{\theta}^t \gets \frac{1}{K} \sum_{k=1}^{K} \theta_k^t$
			\STATE \textbf{Momentum Update:} $\theta_G^t \gets (1-\alpha)\theta_G^{t-1} + \alpha \bar{\theta}^t$
			\ENDFOR
			\STATE \textbf{Return} $\theta_G^T$
		\end{algorithmic}
	\end{algorithm}
	
	This selective and momentum-enhanced aggregation significantly reduces communication overhead by approximately 62\%, as the local encoders remain strictly on the client side, never being transmitted to the server.
	
	\subsection{Local Training and Objective Function}
	Each client $k$ optimizes its local parameters $\theta_k$ by minimizing a multi-objective loss function.
	The total local loss, $\mathcal{L}_{local}$, combines reconstruction accuracy and latent space regularization:
	\begin{equation}
		\mathcal{L}_{local} = \frac{1}{N_k} \sum_{i=1}^{N_k} \sum_{j=1}^{d} (x_{i,j} - \hat{x}_{i,j})^2 + \beta \, \mathcal{D}_{KL}(q_\phi(z|x) \| p(z))
	\end{equation}
	where $N_k$ is the number of local samples for client $k$, and $d$ is the feature dimensionality.
	To ensure numerical precision, the Kullback-Leibler (KL) divergence term is analytically expanded as:
	\begin{equation}
		\mathcal{D}_{KL} = -\frac{1}{2} \sum_{m=1}^{M} \left( 1 + \log(\sigma_m^2) - \mu_m^2 - \sigma_m^2 \right)
	\end{equation}
	where $M$ denotes the dimensionality of the latent space.
	During optimization, each client applies stochastic gradient descent with the AdamW optimizer and enforces a strict gradient clipping threshold $\zeta=0.5$ to maintain numerical stability against noisy or outlier data packets.
	
	The complete procedure for local training, including the computation of $\mathcal{L}_{local}$ and gradient updates, is summarized in Algorithm~\ref{alg:client}.
	
	\begin{algorithm}[htbp]
		\caption{Local Client Optimization Procedure}
		\label{alg:client}
		\begin{algorithmic}[1]
			\STATE \textbf{Input:} Local dataset $\mathcal{D}_k$, Global weights $\theta_G$, Epochs $E$
			\STATE \textbf{Output:} Weight update $\Delta \theta_k$
			\STATE \textbf{Initialize:} $\theta_k \leftarrow \theta_G$
			\FOR{each epoch $e=1$ to $E$}
			\FOR{batch $(x, y) \in \mathcal{D}_k$}
			\STATE Compute conditioned latent vector:
			\[
			z_{\mathrm{cond}} = \text{Algorithm~\ref{alg:acvae}}(x, y)
			\]
			\STATE Reconstruct input: $\hat{x} \leftarrow p_\theta(z_{\mathrm{cond}})$
			\STATE Compute local loss: $$\mathcal{L}_{local} \leftarrow \text{MSE}(x, \hat{x}) + \beta \mathcal{D}_{KL}(\mu, \sigma)$$
			\STATE Compute gradients: $$\nabla \theta_k \leftarrow \text{clip}(\text{Gradients}(\mathcal{L}_{local}), -\zeta, \zeta)$$
			\STATE Update weights: $\theta_k \leftarrow \text{AdamW}(\theta_k, \nabla \theta_k)$
			\ENDFOR
			\ENDFOR
			\STATE \textbf{Return} $\Delta \theta_k = \theta_k - \theta_G$
		\end{algorithmic}
	\end{algorithm}
	
	\subsection{Momentum-Stabilized Aggregation (MSA)}
	\label{sec:msa}
	To mitigate model drift caused by non-IID data, the server implements MSA.
	Upon receiving local updates $\Delta \theta_k^t$, the server first filters potential outliers using an element-wise clamping function:
	\begin{equation}
		\Delta \bar{\theta}_{agg}^t = \frac{1}{|\mathcal{S}_t|} \sum_{k \in \mathcal{S}_t} \text{clamp}(\Delta \theta_k^t, -\tau, \tau)
	\end{equation}
	where $\text{clamp}(\cdot)$ limits each parameter to the range $[-\tau, \tau]$. The hyperparameters $\tau$ (clamping threshold) and $\gamma$ (momentum coefficient) are set as shown in Table~\ref{tab:hyperparams}.
	
	The global model is then updated using a momentum-based velocity $\mathcal{V}$ to smooth the convergence path:
	\begin{equation}
		\mathcal{V}^t = \gamma \Delta \bar{\theta}_{agg}^t + (1-\gamma) \mathcal{V}^{t-1},
	\end{equation}
	\begin{equation}
		\theta_G^{t+1} = \theta_G^t + \eta \, \mathcal{V}^t
	\end{equation}
	where $\eta$ is the global learning rate.
	Algorithmically, MSA is implemented server-side within the Flower aggregation interface.
	Clamped, sample-weighted client updates are accumulated into the momentum buffer $\mathcal{V}$ and applied to the global model parameters at each communication round.
	
	\subsection{Hybrid Latent Feature Classification}
	The final detection phase employs a hybrid strategy where the global F-ACVAE acts as a feature extractor.
	Once federated training converges, the encoder maps raw traffic into the robust, low-dimensional latent space $z$.
	These features are then fed into a random forest (RF) ensemble:
	\begin{equation}
		\hat{y} = \text{argmax}_{c \in \{1,\cdots,C\}} \frac{1}{B} \sum_{b=1}^{B} I(T_b(z) = c)
	\end{equation}
	where $B$ is the number of decision trees.
	This approach maximizes the detection performance by combining generative modeling with discriminative ensemble learning.
	At inference time, the posterior mean of the encoder is used as a deterministic latent representation, which is subsequently classified using a RF ensemble via majority voting.
	
	\section{Experimental Settings}
	\label{sec:experiments}
	
	This section describes the experimental setup of F-ACVAE on the N-BaIoT dataset, using nine device-specific non-IID subsets to simulate realistic botnet traffic conditions.
	We first outline the performance metrics, hyperparameter configurations, dataset characteristics, and implementation details.
	
	\subsection{Performance Metrics}
	We evaluate the proposed F-ACVAE against several state-of-the-art intrusion detection models under identical non-IID settings on the N-BaIoT dataset.
	The baselines include STA, CSAEC, MAE, and MVAE as representative autoencoder-based approaches, along with the centralized constrained twin variational autoencoder (CTVAE) \cite{dinh2024}, which serves as a leading generative model for semi-supervised anomaly detection in IoT environments.
	These benchmarks provide a comprehensive framework to evaluate F-ACVAE's ability to handle data heterogeneity, severe class imbalance, and privacy concerns in decentralized frameworks.
	
	Performance is evaluated using the following standard metrics:
	
	\begin{equation}
		\text{Accuracy} = \frac{\text{TP} + \text{TN}}{\text{TP} + \text{TN} + \text{FP} + \text{FN}},
	\end{equation}
	where TP, TN, FP, and FN denote true positives, true negatives, false positives, and false negatives, respectively.
	\begin{equation}
		\text{Precision} = \frac{\text{TP}}{\text{TP} + \text{FP}},
	\end{equation}
	where precision measures the proportion of correctly identified intrusions among all instances flagged as malicious.
	\begin{equation}
		\text{Recall} = \frac{\text{TP}}{\text{TP} + \text{FN}},
	\end{equation}
	where recall indicates the fraction of actual malicious instances that are correctly detected.
	\begin{equation}
		\text{Macro F1-score} = \frac{1}{C} \sum_{i=1}^{C} 2 \times \frac{\text{Precision}_i \times \text{Recall}_i}{\text{Precision}_i + \text{Recall}_i},
	\end{equation}
	where $C$ is the number of classes.
	Being macro-averaged, this metric assigns equal importance to the benign class and all individual attack types, making it the primary evaluation measure in this study due to the extreme class imbalance observed in real-world IoT botnet traffic.
	
	\subsection{Parameter Settings}
	To configure the federated training process, we simulate a decentralized environment with $K = 9$ device-specific clients participating in $T = 10$ communication rounds.
	In each round, every client performs $E = 5$ local training epochs using a mini-batch size of $B = 128$.
	
	The global F-ACVAE model adopts an encoder--decoder architecture with three hidden layers of sizes $[128, 64, 32]$.
	All networks employ the LeakyReLU activation function with a negative slope of $0.2$.
	Model optimization is carried out using the AdamW optimizer with a learning rate of $\eta = 2 \times 10^{-4}$ and a weight decay coefficient of $\lambda_w = 5 \times 10^{-6}$.
	
	The input feature dimension is fixed to $d_x = 115$, corresponding to the N-BaIoT dataset, while the latent space dimensionality is set to $d_z = 10$.
	To ensure stable variational inference and prevent posterior collapse under highly imbalanced non-IID data, the KL-divergence regularization weight is set to $\beta = 0.3$.
	The class-conditioning scale is fixed to $\lambda_c = 1.0$ to balance discriminative guidance and generative flexibility.
	
	For the proposed CMGA strategy, the aggregation momentum is set to $\alpha = 0.1$, and the momentum coefficient is fixed at $\gamma = 0.1$ to smooth global updates across heterogeneous clients.
	Additionally, parameter updates are constrained using a clamping threshold of $\tau = 0.2$ to mitigate abrupt shifts caused by severe label skew.
	
	All hyperparameter configurations are summarized in Table~\ref{tab:hyperparams}.
	
	\begin{table}[htbp]
		\centering
		\caption{Hyperparameter configurations for F-ACVAE}
		\label{tab:hyperparams}
		\begin{tabular}{lcc}
			\toprule
			\textbf{Parameter}          & \textbf{Symbol}   & \textbf{Value}        \\
			\midrule
			Total number of clients     & $K$               & 9                     \\
			Federated rounds            & $T$               & 10                    \\
			Local epochs                & $E$               & 5                     \\
			Batch size                  & $B$               & 128                   \\
			Learning rate               & $\eta$            & $2 \times 10^{-4}$    \\
			Optimizer                   & --                & AdamW                 \\
			Weight decay                & $\lambda_w$       & $5 \times 10^{-6}$    \\
			Input dimension             & $d_x$             & 115                   \\
			Latent dimension            & $d_z$             & 10                    \\
			KL-divergence weight        & $\beta$           & 0.3                   \\
			Class-conditioning scale    & $\lambda_c$       & 1.0                   \\
			Aggregation momentum        & $\alpha$          & 0.1                   \\
			Momentum coefficient        & $\gamma$          & 0.1                   \\
			Clamping threshold          & $\tau$            & 0.2                   \\
			Hidden layers               & --                & [128, 64, 32]         \\
			Activation function         & --                & LeakyReLU (0.2)       \\
			\bottomrule
		\end{tabular}
	\end{table}
	
	\subsection{Datasets}
	F-ACVAE is evaluated on the N-BaIoT dataset~\cite{meidan2018nbaiot}\footnote{The N-BaIoT dataset is publicly available at \url{https://archive.ics.uci.edu/dataset/442/detection_of_iot_botnet_attacks_n_baiot}.}, which captures network traffic from nine commercial IoT devices infected by the Mirai and Bashlite (Gafgyt) botnets. Table~\ref{tab:nbaiot-summary} summarizes key dataset characteristics, including total instances, feature dimensionality, class distribution, and attack types. Table~\ref{tab:nbaiot-attack-types} details the attack vectors, which categorizes them by botnet family and provides a brief description of each attack mechanism.
	
	\begin{table}[htbp]
		\centering
		\caption{Summary of the N-BaIoT dataset}
		\label{tab:nbaiot-summary}
		\renewcommand{\arraystretch}{1.2}
		\begin{tabular}{lr}
			\toprule
			\textbf{Property}       & \textbf{Value}            \\
			\midrule
			Number of IoT devices   & 9                         \\
			Total instances         & 7,062,606                 \\
			Number of features      & 115                       \\
			Benign instances        & 555,932                   \\
			Malicious instances     & 6,506,674                 \\
			Attack Families         & 2 (Mirai, Gafgyt)         \\
			Distinct Attack Vectors & 10                        \\
			Train / Test split      & 70\% / 30\% (Stratified)  \\
			\bottomrule
		\end{tabular}
	\end{table}
	
	\begin{table}[htbp]
		\centering
		\caption{Attack types in the N-BaIoT dataset}
		\label{tab:nbaiot-attack-types}
		\begin{tabularx}{\columnwidth}{l l X}
			\toprule
			\textbf{\shortstack{Botnet \\ Family}} & \textbf{\shortstack{Attack \\ Type}} & \textbf{Description} \\
			\midrule
			\multirow{5}{*}{\shortstack{Bashlite \\ (Gafgyt)}}
			& Scan     & Horizontal scanning for vulnerable devices \\
			& Junk     & Sending spam-like packets \\
			& TCP      & TCP flood to exhaust resources \\
			& UDP      & UDP flood to overwhelm the target \\
			& UDPplain & Plain UDP flood without payload tricks \\
			\midrule
			\multirow{5}{*}{\shortstack{Mirai}}
			& Ack      & ACK flood to consume bandwidth/resources \\
			& Scan     & Aggressive port scanning \\
			& Syn      & SYN flood (TCP handshake attack) \\
			& UDP      & Massive UDP flood saturating network \\
			& UDPplain & Simple UDP flood with generic payloads \\
			\bottomrule
		\end{tabularx}
	\end{table}
	
	To simulate realistic non-IID conditions, we utilize nine device-specific subsets, as summarized in Table~\ref{tab:datasets}. 
	These subsets exhibit data heterogeneity, including label skew and feature distribution shifts due to device-specific traffic patterns, effectively representing the challenges of decentralized IoT networks for both binary and multiclass classification tasks.
	
	\begin{table*}[htbp]
		\centering
		\caption{Device-specific subsets of the N-BaIoT dataset}
		\label{tab:datasets}
		\begin{tabular*}{\textwidth}{@{\extracolsep{\fill}}llllccccc}
			\toprule
			\textbf{ID} & \textbf{Code} & \textbf{Device Model} & \textbf{Category} & \textbf{Protocols} & \textbf{Attack Families} & \textbf{Attack Ratio} & \textbf{Classes} & \textbf{Instances} \\ 
			\midrule
			IoT-01 & DanG-6   & Danmini            & Doorbell      & TCP/UDP & Gafgyt          & 88.5\% & 6  & 1,018,298 \\
			IoT-02 & PhiG-6   & Philips B120N/10   & Baby Monitor  & TCP/UDP & Gafgyt          & 91.2\% & 6  & 1,091,330 \\
			IoT-03 & 838G-6   & Provision PT-838TS & IP Camera     & TCP/UDP & Gafgyt          & 85.0\% & 6  & 835,656   \\
			IoT-04 & EcoG-6   & Ecobee             & Thermostat    & TCP/UDP & Gafgyt          & 82.3\% & 6  & 724,321   \\
			IoT-05 & 737G-6   & Provision PT-737E  & IP Camera     & TCP/UDP & Gafgyt          & 86.1\% & 6  & 828,244   \\
			IoT-06 & EcoMG-11 & Ecobee             & Thermostat    & TCP/UDP & Mirai \& Gafgyt & 94.7\% & 11 & 1,234,550 \\
			IoT-07 & 838MG-11 & Provision PT-838TS & IP Camera     & TCP/UDP & Mirai \& Gafgyt & 93.2\% & 11 & 1,156,000 \\
			IoT-08 & 737GUC-2 & Provision PT-737E  & IP Camera     & UDP     & Gafgyt          & 78.9\% & 2  & 450,120   \\
			IoT-09 & 838GUC-2 & Provision PT-838TS & IP Camera     & UDP     & Gafgyt          & 79.5\% & 2  & 482,000   \\ 
			\bottomrule
		\end{tabular*}
	\end{table*}
	
	For fair comparison with the centralized CTVAE baseline~\cite{dinh2024}, we employ a standard 70\%/30\% train-test split using stratified sampling to preserve class proportions within each subset.
	
	\subsection{Implementation and Reproducibility}
	F-ACVAE is implemented in PyTorch 2.1 and the Flower 1.8 federated learning library.
	To ensure reproducibility, the complete source code, detailed instructions, and pre-trained configurations are publicly available at: \url{https://github.com/mohamad-ansarimehr/F-ACVAE}.
	Additionally, the repository is archived for long-term accessibility with DOI: \url{https://doi.org/10.5281/zenodo.17919997}.
	
	\section{Results and Analysis}
	\label{sec:results}
	
	This section presents the experimental results, highlighting the strong performance of F-ACVAE compared with state-of-the-art baselines.
	The proposed framework achieves an accuracy and macro F1-score of 99\% while reducing communication overhead by approximately 62\%.
	To understand why the proposed F-ACVAE is better, we analyze four key aspects: the advantages of decentralized design by using federated learning, the resulting performance improvements, enhanced privacy preservation, and communication efficiency powered by the custom Flower strategy (CMGA).
	
	\subsection{Decentralization Advantages}
	F-ACVAE benefits from the federated paradigm combined with CMGA in three ways:
	
	\begin{enumerate}
		\item Implicit regularization: Centralized training on the full dataset often overfits to noise and outliers.
		By averaging updates from heterogeneous client data (Algorithm~\ref{alg:client}), the federated approach produces more robust and generalizable latent representations.
		\item Mitigation of latent space collapse: Centralized VAEs can suffer latent space collapse under highly imbalanced data.
		Momentum-stabilized aggregation (MSA; see Section~\ref{sec:msa}) stabilizes the shared components, maintaining a well-structured latent space across clients.
		\item Enhanced robustness to heterogeneity: The momentum mechanism in CMGA reduces sudden changes in global updates caused by severe label imbalance or feature distribution shifts, ensuring the shared decoder learns a stable reconstruction mapping.
	\end{enumerate}
	
\begin{figure}[htbp]
		\centering
		\begin{tikzpicture}
			\begin{axis}[
				width=\columnwidth,
				height=6.5cm,
				xlabel={Federated Rounds ($T$)},
				ylabel={Performance (\%)},
				xmin=0.5, xmax=10.5,
				ymin=50, ymax=100,
				xtick={1,2,3,4,5,6,7,8,9,10},
				ytick={50,60,70,80,90,100},
				legend pos=north west,
				legend style={font=\small},
				ymajorgrids=true,
				grid style=dashed,
				tick label style={font=\small},
				label style={font=\small},
				]
				
				\addplot[
				color=blue,
				mark=square,
				very thick
				]
				coordinates {
					(1,57.3)(2,57.3)(3,57.3)(4,62.1)(5,67.2)
					(6,81.9)(7,94.1)(8,98.8)(9,98.8)(10,99.7)
				};
				\addlegendentry{Accuracy}
				
				\addplot[
				color=red,
				mark=circle,
				very thick
				]
				coordinates {
					(1,72.9)(2,72.9)(3,72.9)(4,74.4)(5,76.6)
					(6,85.3)(7,94.8)(8,99.0)(9,99.0)(10,99.7)
				};
				\addlegendentry{Macro F1-score}
				
			\end{axis}
		\end{tikzpicture}
		\caption{Convergence of F-ACVAE on highly non-IID N-BaIoT data (IoT-01 subset, representative example).
			Both accuracy and macro F1-score exceed 99\% within 10 federated rounds ($T=10$).}
		\label{fig:ablation-rounds}
	\end{figure}
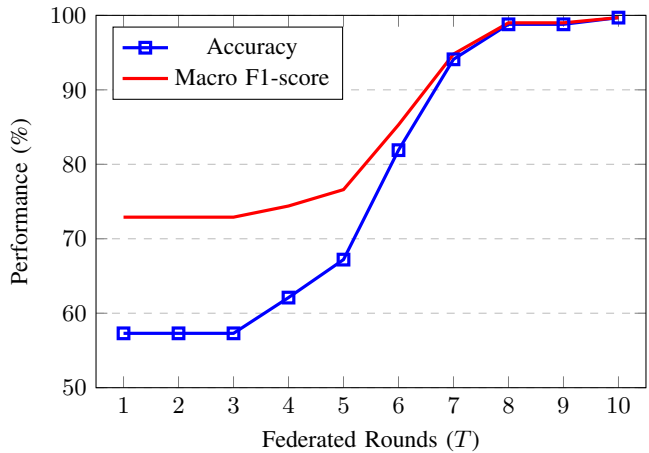

	As shown in Figure~\ref{fig:ablation-rounds}, we analyze the convergence behavior of F-ACVAE on the IoT-01 subset of the highly non-IID N-BaIoT dataset, which serves as a representative case for evaluating the framework's performance.
	The model is trained over $T=10$ federated communication rounds to examine stability and convergence under the selective aggregation mechanism implemented via CMGA (see Algorithm~\ref{alg:flower-cmga}).
	
	Results show that F-ACVAE rapidly converges within a few communication rounds, achieving both accuracy and macro F1-score of 99\% without oscillations or instability.
	These observations confirm that the proposed aggregation strategy effectively stabilizes global model updates across heterogeneous clients, even under extreme non-IID data distributions.
	
	\subsection{Performance Benefits}
	Tables~\ref{tab:accuracy-comparison} and~\ref{tab:f1-comparison} present the performance of F-ACVAE compared to state-of-the-art baselines across all nine device-specific subsets of the N-BaIoT dataset.
	F-ACVAE achieves the highest accuracy and macro F1-score of 99\% across all evaluated scenarios.
	
	Notably, F-ACVAE outperforms the centralized CTVAE \cite{dinh2024} in most subsets, highlighting the advantages of the proposed federated training framework even when compared with advanced generative models.
	This performance gain is attributed to the selective parameter aggregation mechanism and the CMGA strategy, which effectively mitigate extreme non-IID challenges while preserving privacy.
	
	Overall, by maintaining a well-conditioned latent structure without raw data exchange, F-ACVAE emerges as a robust, efficient, and privacy-preserving solution for intrusion detection in highly heterogeneous IoT environments.
	
	\begin{table}[htbp]
		\centering
		\caption{Accuracy (\%) comparison}
		\label{tab:accuracy-comparison}
		\resizebox{\columnwidth}{!}{
			\begin{tabular}{l cccccc}
				\toprule
				\textbf{Dataset} & \textbf{STA} & \textbf{CSAEC} & \textbf{MAE} & \textbf{MVAE} & \textbf{CTVAE} & \textbf{F-ACVAE} \\
				\midrule
				IoT-01 & 67.2 & 92.0 & 92.1 & 67.2 & 93.0 & \textbf{99.7} \\
				IoT-02 & 75.2 & 94.2 & 94.1 & 74.2 & 95.0 & \textbf{99.1} \\
				IoT-03 & 70.9 & 92.8 & 93.4 & 70.5 & 93.9 & \textbf{98.1} \\
				IoT-04 & 61.3 & 90.6 & 90.9 & 60.0 & 91.8 & \textbf{99.3} \\
				IoT-05 & 65.6 & 92.1 & 92.2 & 65.7 & 93.1 & \textbf{97.9} \\
				IoT-06 & 85.0 & 92.3 & 78.1 & 62.8 & 94.5 & \textbf{98.9} \\
				IoT-07 & 85.8 & 96.5 & 94.2 & 76.8 & 96.6 & \textbf{99.6} \\
				IoT-08 & 97.6 & 90.0 & 95.5 & 88.5 & \textbf{100.0} & 98.7 \\
				IoT-09 & 95.3 & 80.2 & 94.8 & 90.1 & 98.9 & \textbf{99.5} \\
				\midrule
				\textbf{Average} & 78.2 & 91.2 & 91.7 & 72.9 & 95.2 & \textbf{99.0} \\
				\bottomrule
			\end{tabular}
		}
	\end{table}
	
	\begin{table}[htbp]
		\centering
		\caption{Macro F1-score (\%) comparison}
		\label{tab:f1-comparison}
		\resizebox{\columnwidth}{!}{
			\begin{tabular}{l cccccc}
				\toprule
				\textbf{Dataset} & \textbf{STA} & \textbf{CSAEC} & \textbf{MAE} & \textbf{MVAE} & \textbf{CTVAE} & \textbf{F-ACVAE} \\
				\midrule
				IoT-01 & 55.7 & 88.9 & 89.2 & 64.9 & 90.9 & \textbf{99.7} \\
				IoT-02 & 66.4 & 91.8 & 91.9 & 71.5 & 93.5 & \textbf{99.4} \\
				IoT-03 & 60.4 & 89.9 & 91.3 & 67.9 & 92.2 & \textbf{98.5} \\
				IoT-04 & 47.7 & 87.0 & 87.6 & 57.0 & 89.5 & \textbf{98.8} \\
				IoT-05 & 53.6 & 89.0 & 89.3 & 62.8 & 91.1 & \textbf{98.0} \\
				IoT-06 & 79.7 & 91.8 & 76.9 & 61.7 & 93.3 & \textbf{98.6} \\
				IoT-07 & 80.7 & 95.1 & 94.2 & 75.5 & 95.5 & \textbf{99.6} \\
				IoT-08 & 97.9 & 92.8 & 96.4 & 91.9 & \textbf{100.0} & 98.9 \\
				IoT-09 & 99.7 & 85.7 & 95.7 & 92.2 & 99.0 & \textbf{99.8} \\
				\midrule
				\textbf{Average} & 70.9 & 90.2 & 90.3 & 71.7 & 93.9 & \textbf{99.0} \\
				\bottomrule
			\end{tabular}
		}
	\end{table}
	
		
	
	\subsection{Preserving Privacy}
	F-ACVAE preserves privacy by keeping encoders local on each client and aggregating only the shared mapper and decoder (Algorithm~\ref{alg:acvae}).
	As a result, raw data never leaves the device and latent representations remain local, providing strong protection against inversion and membership inference attacks.
	
	\subsection{Communication Efficiency}
	F-ACVAE leverages CMGA (Algorithm~\ref{alg:flower-cmga}) to transmit only a subset of model weights, while keeping the model structure (mapper, decoder, class-specific Gaussian priors) fixed, significantly reducing communication overhead.  
	Let $r_\text{sent}$ denote the fraction of model weights sent in each round:
	
	\begin{equation}
		r_\text{sent} = \frac{W_\text{sent}}{W_\text{total}} \approx 0.38
		\quad \Rightarrow \quad
		r_\text{reduced} = 1 - r_\text{sent} \approx 0.62
	\end{equation}
	
	In practice, only about 38\% of the weights are transmitted per round, which reduces communication traffic by approximately 62\%.
	Stable updates are maintained through the momentum-stabilized aggregation mechanism (MSA; see Section~\ref{sec:msa}), making the framework suitable for resource-constrained IoT gateways.
	
		%
		%
		%
		%
	\section{Future Work}
	\label{sec:future}
	
	The success of F-ACVAE demonstrates its ability to preserve discriminative latent structures under federated learning. 
	This opens several directions for future research, including algorithmic optimization, enhanced security, and validation in real-world IoT deployments.
	
	\subsection{Algorithm and Architecture Enhancements}
	Future work will focus on improving the CMGA strategy, including adaptive client selection based on data quality or gradient similarity to mitigate label skew. 
	We also plan to explore privacy-preserving knowledge transfer from local encoders to the global model, for example through federated distillation, to enhance feature extraction without compromising privacy. 
	Extending F-ACVAE to fully unsupervised settings will remove the need for attack labels, moving toward practical zero-shot anomaly detection. 
	Finally, current assumptions such as a fixed number of clients may be relaxed to handle highly dynamic IoT networks, where adaptive tuning of proximal constraints is required to maintain convergence stability under client churn.
	
	\subsection{Security and Resilience}
	Although F-ACVAE secures data privacy through federated learning, further work is needed to strengthen protection against federated threats. 
	This includes robust defenses against model poisoning (Byzantine resilience) and inference attacks such as membership inference targeting the global model. 
	Incorporating cryptographic techniques, such as secure multi-party computation (SMPC) and differential privacy, will be essential to ensure the integrity of the CMGA process.
	
	\subsection{Scalability and Real-World Deployment}
	Future evaluations should examine F-ACVAE in complex real-world scenarios, assessing scalability, latency, and energy efficiency on resource-constrained devices such as Raspberry Pi. 
	Extending its application to broader IoT domains—including industrial IoT, smart health, and critical infrastructure—is critical. 
	Additionally, evaluation with complex multimodal datasets will be essential to confirm the generalized effectiveness of the constrained latent space approach.
	
	\section{Conclusion}
	\label{sec:conclusion}
	
	In this paper, we introduced F-ACVAE, a robust federated learning framework designed to address the challenges of privacy preservation and data heterogeneity in IoT intrusion detection.
	By integrating adaptive conditional variational autoencoding with the proposed CMGA strategy, the framework effectively separates local feature extraction from global generative modeling.
	This architectural design, combined with momentum-based updates, effectively mitigates weight drift and prevents latent space collapse in non-IID scenarios.
	
	Our experimental evaluation on the N-BaIoT dataset demonstrates that F-ACVAE achieves accuracy and macro F1-score of 99\%.
	Furthermore, the selective aggregation mechanism reduces communication overhead by approximately 62\%, making it highly suitable for resource-constrained IoT devices.
	These results demonstrate that F-ACVAE achieves both high detection performance and strong privacy preservation, making it a robust solution for decentralized IoT security ecosystems.
	
	Future work will focus on advancing the CMGA strategy through adaptive client selection and privacy-preserving knowledge transfer.
	It will also enhance defenses against federated threats by employing techniques such as SMPC and differential privacy.
	Finally, we will evaluate the performance of F-ACVAE across diverse IoT domains and resource-constrained hardware platforms.
	
	\bibliographystyle{IEEEtran}

\end{document}